\documentclass[10pt,twocolumn,letterpaper]{article}

\usepackage{cvpr}              %

\pdfoutput=1

\usepackage[dvipsnames]{xcolor}

\definecolor{cvprblue}{rgb}{0.21,0.49,0.74}
\usepackage[pagebackref,breaklinks,colorlinks,citecolor=cvprblue]{hyperref}
\usepackage{amsmath}

\newcommand{\vect}[1]{\boldsymbol{#1}}

\def\paperID{5129} %
\def\confName{CVPR}
\def\confYear{2024}

\title{Video Interpolation with Diffusion Models}

\newcommand{\footremember}[2]{%
   \thanks{#2}
    \newcounter{#1}
    \setcounter{#1}{\value{footnote}}%
}
\newcommand{\footrecall}[1]{%
    \footnotemark[\value{#1}]%
} 
\author{Siddhant Jain\footremember{firstauthor}{Equal contribution.}\\
Google Research\\
{\tt\small siddhantjain@google.com}
\and
Daniel Watson\footrecall{firstauthor}\\
Google DeepMind\\
{\tt\small watsondaniel@google.com}
\and
Eric Tabellion\\
Google Research\\
{\tt\small etabellion@google.com}
\and
Aleksander Hołyński\\
Google Research\\
{\tt\small holynski@google.com}
\and 
Ben Poole\\
Google DeepMind\\
{\tt\small pooleb@google.com}
\and
Janne Kontkanen\\
Google Research\\
{\tt\small jkontkanen@google.com}
}

\begin{document}
\maketitle
\begin{figure*}[t!]
    \centering
    \includegraphics[width=\textwidth]{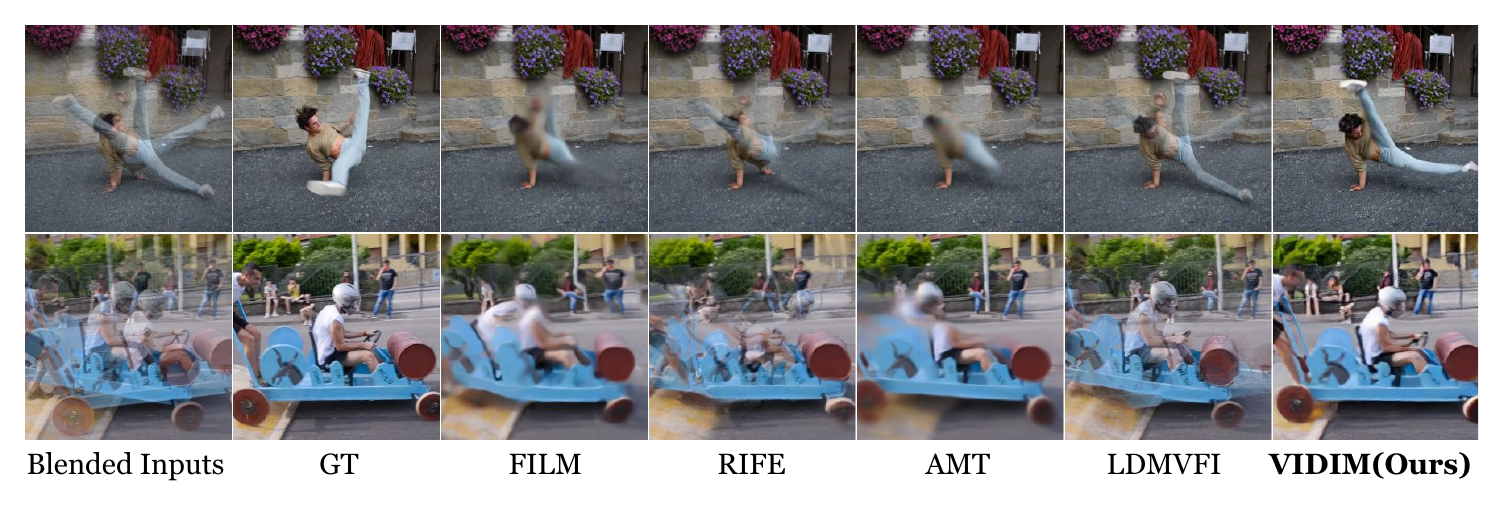}
    \caption{Frame interpolation for very large and ambiguous motion.  
    The middle frame of an interpolated video with FILM\citep{reda2022film}, RIFE ~\citep{rife}, LDMVFI ~\citep{danier2023ldmvfi} and AMT\citep{amt} shows large blurry artifacts. VIDIM, however, is able to recover a plausible output frame. Note that due to the ambiguity of the problem, VIDIM's output is not always similar to the ground truth (especially clear in the top example), but corresponds to a different choice of motion. See the \href{https://vidim-interpolation.github.io/}{Supplementary Website} for video outputs.}
    \label{fig:teaser}
    \vspace{-2ex}
\end{figure*}

\begin{abstract}
We present VIDIM, a generative model for video interpolation, which creates short videos given a start and end frame. In order to achieve high fidelity and generate motions unseen in the input data, VIDIM uses cascaded diffusion models to first generate the target video at low resolution, and then generate the high-resolution video conditioned on the low-resolution generated video. We compare VIDIM to previous state-of-the-art methods on video interpolation, and demonstrate how such works fail in most settings where the underlying motion is complex, nonlinear, or ambiguous while VIDIM can easily handle such cases. We additionally demonstrate how classifier-free guidance on the start and end frame and conditioning the super-resolution model on the original high-resolution frames without additional parameters unlocks high-fidelity results. VIDIM is fast to sample from as it jointly denoises all the frames to be generated, requires less than a billion parameters per diffusion model to produce compelling results, and still enjoys scalability and improved quality at larger parameter counts. Please see our project page at \href{https://vidim-interpolation.github.io/}{vidim-interpolation.github.io}.
\end{abstract}
    
\section{Introduction}
\label{sec:intro}
Diffusion Models \citep{sohl2015deep,song2019generative,ho2020denoising} have recently exploded in popularity for generative modeling of images and other forms of continuous data such as audio \citep{chen2020wavegrad} and video \citep{ho2022video}. Compared to previous methods for generative modeling such as Generative Adversarial Networks (GANs) \citep{goodfellow2020generative}, diffusion models enjoy significantly more training stability due to the fact that they optimize the evidence lower bound (ELBO) \citep{kingma2013auto}, as opposed to having the complex dynamics of two models in a zero-sum game like GANs, and do not suffer from posterior collapse like variational autoencoders (VAEs) due to the limited form of the approximate posterior. This ease of training has led to significant advances in generative modeling, such as compelling results in text-to-image \citep{ramesh2022hierarchical,saharia2022photorealistic_imagen_video,rombach2022high} and text-to-video \citep{villegas2022phenaki,ho2022cascaded_imagen} generation, image-conditioned generation tasks \citep{saharia2022palette}, and even 3D novel view synthesis \citep{poole2022dreamfusion,liu2023zero,watson2022novel}.

In this paper, we explore the specific task of video interpolation with diffusion models.
Video interpolation refers to the problem of generating intermediate frames between two consecutive frames of video. Video interpolation techniques have been used for many desirable applications, e.g., generating slow motion videos from existing videos, video frame rate up-sampling (\eg 30 fps $\rightarrow$ 60 fps) or interpolating between near-duplicate photographs.

Numerous methods have been proposed in prior work~\citep{dong2023vfisurvey}, but even the state-of-the-art~\citep{amt,rife,reda2022film} fails to generate plausible interpolations when the start and end frame become increasingly distinct, as these methods rely on linear or unambiguous motion. Moreover, while existing video diffusion models can be used for the video interpolation task \citep{ho2022video,ho2022cascaded_imagen}, as we carefully study in this work, quantitative and qualitative results improve significantly by explicitly training generative models that are conditioned on the start and end frames, as generative models have the key advantage of producing samples, as opposed to predicting the mean. As we will show, this is also strongly supported by ratings from human observers.

In this work, we show that diffusion based generative models can overcome the limitations of prior state-of-the-art models for video interpolation. We summarize our main contributions below:
\begin{itemize}
    \item We develop a cascaded video interpolation diffusion model, which we dub \textbf{VIDIM}, capable of generating high-quality videos in between two input frames.
    \item We carefully ablate some of the design choices of VIDIM, including parameter sharing to process conditioning frames and the use of classifier-free guidance \citep{ho2022classifier}, demonstrating their importance to achieve good results.
    \item We propose two curated difficult datasets targeted for generative frame interpolation: Davis-7 and UCF101-7, based on widely used Davis~\cite{davis} and UCF101~\cite{soomro2012ucf101} datasets. 
    \item We show that VIDIM generally achieves better results compared to prior state-of-the-art in these difficult interpolation problems across generative modeling metrics. We show by user study that VIDIM is almost always preferred over the baselines in qualitative evaluation.
\end{itemize}

\section{Background and Related Work}
\label{sec:background}

\label{sec:related_video_frame_interpolation}
Video frame interpolation (VFI) is a classic computer vision problem with a sizable body of existing work. VFI is closely related to optical flow computation, which is an equally deeply studied problem. Instead of attempting a comprehensive list of works in these areas we 
discuss the recent state-of-the-art that is most relevant for our work. For a recent survey on this topic, see~\cite{dong2023vfisurvey}.

Most recent video frame interpolation architectures contain a feature extractor (e.g. decoder) correspondence estimation and image warping (e.g. optical flow) and frame synthesis (e.g. decoder). Most works also agree that optical flow is best learned for the frame interpolation task specifically~\cite{vimeo, reda2022film, kong2022irfnet,danier2022stmfnet} or fine tuned for it~\cite{park2021ABME, niklaus2020softsplat, jiang2017superslomo}. Some methods use backward warping (gather)~\cite{reda2022film, park2021ABME, kong2022irfnet, rife, amt} while others use forward warping (splatting/scatter)~\cite{niklaus2020softsplat}.

Recent works employ hybrid CNN and transformer architectures~\cite{zhang2023emavfi,lu2022vfiformer} and all-pairs dense feature matching~\cite{amt} inspired by the state-of-the-art optical flow method~\cite{teed2020raft}. Regardless of the details, the progress of video frame interpolators have been driven by benchmarks~\cite{vimeo,baker2007middlebury,butler2012sintel,soomro2012ucf101,davis} that are usually prepared to assume linear or mostly unambiguous motion, by either having explicit linearity constraints~\cite{vimeo} or just using samples that are not very far apart in time.
It is rare to go beyond the non-linear motion assumption, although a few methods do assume a quadratic motion model~\cite{liu2019qvi,liu2020eqvi}. Some recent works specialize on large motions~\cite{sim2021xvfi, reda2022film}, but yet only to an extent where the motion is mostly unambiguous and thus can be solved with a non-generative model. In our work we show that when the input images are much further than 1/30s apart, the problem becomes highly ambiguous and best addressed as a conditional generative problem.

LDMVFI \cite{danier2023ldmvfi} and MCVD \cite{voleti2022mcvd} are some recent diffusion based frame interpolatation methods. Different from LDMVFI we model in pixel space and generate the entire video at once which is key for consistent motion. We also focus on explicitly training for video frame interpolation where as MCVD studies a range of video generative modeling tasks.  While it is difficult to design truly fair comparisons as existing video models are often too large and source code is not available \citep{ho2022cascaded_imagen,singer2022make}, the conditioning mechanism can still be studied. \citet{ho2022video} propose the use of imputation (and additionally with a mechanism resembling classifier-free guidance) to adapt video models to be conditioned on input frames. We train a super-resolution model adopting this strategy, and in section \ref{subsec:ablation_highresframe} demonstrate that this approach performs worse despite being having the exact same hyperparameters otherwise.

\label{sec:related_generative_video_models}

\section{Methodology}

We now present the technical details behind VIDIM, our cascaded diffusion model for video interpolation. Prior work has shown that diffusion models do not achieve good sample quality for high-resolution generation with a single model without revising several hyperparameters and architecture details \citep{hoogeboom2023simple}. We in fact tried training a base VIDIM model to generate a 9x128x128\footnote{Not that large resolution by video interpolation standards} video following the changes proposed by \citet{hoogeboom2023simple}, but we found early on that these models did a very poor job at modeling high-frequency details at this resolution. We thus followed the cascaded model strategy of \citet{ho2022cascaded}, i.e., training separate base and super-resolution models. While there is additional overhead to maintaining multiple diffusion models, this still avoids several complexities of latent diffusion models \citep{rombach2022high}, such as finding an optimal encoder-decoder model, and having to address temporal inconsistency in the decoder with other training or fine-tuning procedures \citep{blattmann2023align}. We train two video diffusion models: first we train a \textbf{base model} that is conditioned on 2 64x64 frames and generates 7 64x64 in-between frames. Then we train a \textbf{super-resolution model} conditioned on 2 256x256 frames and 7 64x64 frames that generates the 7 corresponding 256x256 frames. We intentionally chose an odd number of frames to allow evaluating the middle frame, similarly to prior work on video interpolation.

\subsection{Model architecture}
\label{subsec:model_architecture}
VIDIM is inspired by Imagen Video \citep{ho2022cascaded_imagen}, where the UNet architecture \citep{ronneberger2015u} is adapted for video generation by sharing all convolution and self-attention blocks over frames, and feature maps are only allowed to mix over frames with the addition of temporal attention blocks where the query-key-value sequence lengths are the number of frames. Simple positional encodings (differing over frames) for video timestamps normalized to $[0,1]$ are summed to the usual noise level embeddings (identical for all noisy frames). We propagate these embeddings to each UNet block using FiLM \citep{perez2018film}, similarly to \citet{nichol2021improved}. We do the same for the super-resolution model to condition on the high-resolution start and end frames. The super-resolution model only differs from the base model in that (1) it concatenates each (naively upsampled) low-resolution conditioning frame to the noisy high-resolution frames along the \textit{channel} axis, and (2) it downsamples \textit{before} the first convolutional residual block, following \citet{saharia2022photorealistic_imagen_video} to reduce memory usage. For more stable and efficient training, we additionally use attention blocks following \citet{dehghani2023scaling}, which employ query-key normalization and an MLP block that runs in parallel to the attention block.

\paragraph{Parameter-free frame conditioning.} One key innovation we highlight is introducing frame-conditioning without any additional parameters: in both the base and super-resolution models, we condition on the start and end frames simply by feeding these two additional frames to the entire UNet (i.e., concatenating along the \textit{frame} axis). Because the feature maps for each frame additionally depends on the noise levels, we simply set fake noise levels for the conditioning frames as the minimum noise level (maximum log-signal-to-noise-ratio). This adds two new sequence elements to the temporal attention layer, which lets information from the conditioning frames propagate to the rest of the network without any additional parameters. This contrasts other diffusion architectures \citep{zhu2023tryondiffusion,liu2023zero}, where the usual choice is additional cross-attention layers which doesn't scale to condition on more frames and make the parameter count dependent on the number of frames. Other work has found that concatenating extra frames along the \textit{channel} axis following, e.g., \citet{saharia2022palette}, leads to worse sample quality when the generated and conditioning frames are not perfectly aligned \citep{watson2022novel}.

\paragraph{Guidance on conditioning frames.} In our results, we show that classifier-free guidance (CFG) \citep{ho2022classifier} on the conditioning frames is essential to achieve the best sample quality. Similarly to our parameter-free frame conditioning strategy, we would like masked conditioning frames for CFG to play naturally with  parameter sharing across frames. To achieve this, instead of zeroing-out the conditioning frames, we replace them with isotropic Gaussian noise and set their corresponding noise levels to the maximum value. It would also be confounding if we zero-out the timestamps for the conditioning frames, so we instead replace them with a learned null token.

\begin{figure*}[p]
    \centering
    \includegraphics[width=0.80\textwidth]{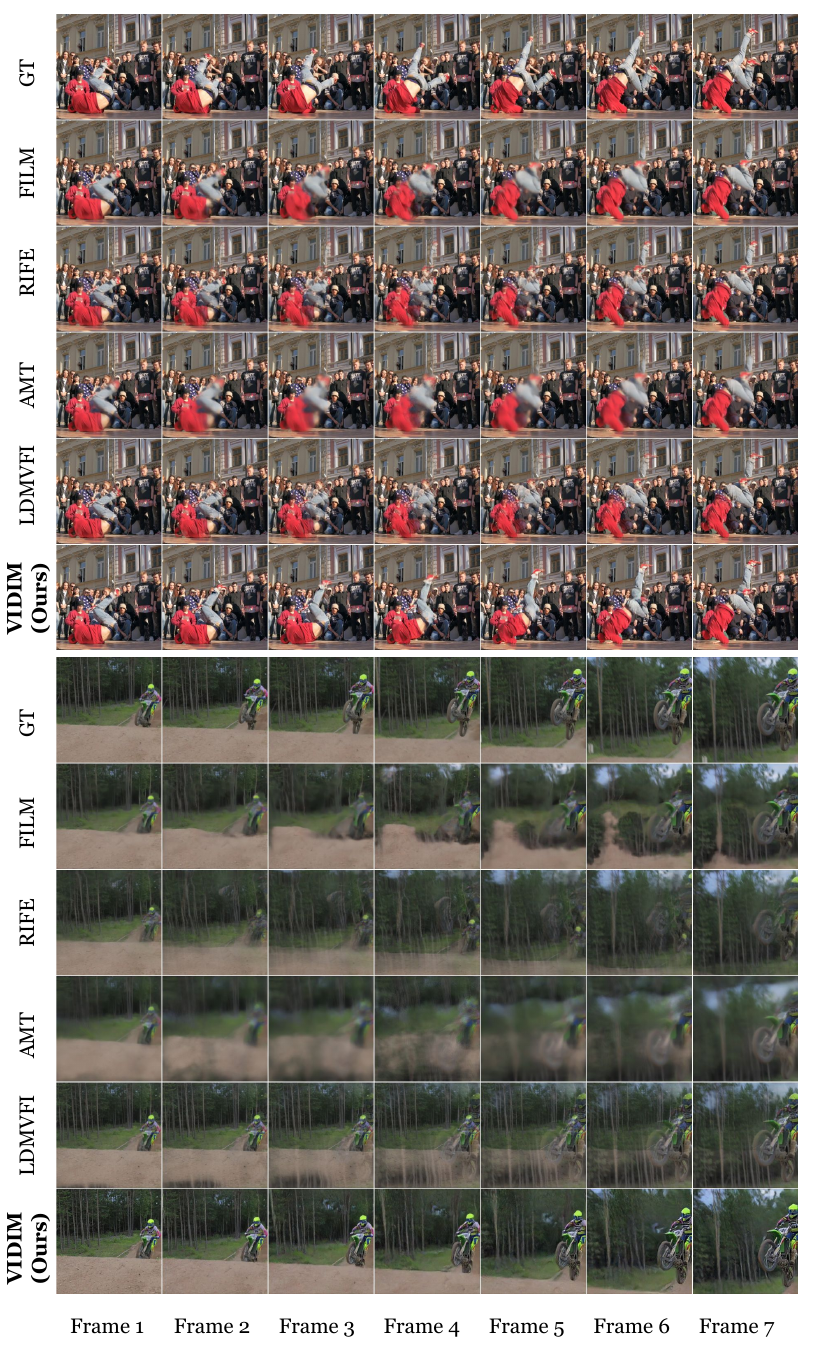}
    \caption{Two examples from DAVIS-9 dataset, showing the predicted in-between frames. {\bf Top}: The break-dancer example demonstrates highly ambiguous motion. Our method can produce plausible video with sharp details whereas the baselines~\cite{reda2022film,rife,amt} trained with regression objective resort into predict blurry images. {\bf Bottom}: On a very large motion with significant perspective change on the dirt bike, the baselines fail to reconstruct sharp results, where as our method produces sharp results with plausible motion.}
    \label{fig:teaser}
    \vspace{4ex}
\end{figure*}

\subsection{Diffusion modeling choices}

\begin{table*}[t]
\small
\renewcommand\tabcolsep{5.0pt}
\centering
\begin{tabular}{ccccc@{\hskip 0.1cm}|cccc}
&  \multicolumn{4}{c}{\textbf{Davis-7 (mid-frame)}}{\hskip 0.1cm} & \multicolumn{4}{c}{\textbf{UCF101-7 (mid-frame)}}  \\
\cline{2-9}
  & PSNR$\uparrow$ & SSIM$\uparrow$ & LPIPS$\downarrow$ & FID$\downarrow$ & PSNR$\uparrow$ & SSIM$\uparrow$ & LPIPS$\downarrow$ & FID$\downarrow$ \\ 
\hline
AMT \citep{amt} & \textbf{20.12} & \textbf{0.4853} & 0.2865 & 69.34 & \textbf{25.16} & \textbf{0.7903} & 0.1691 & 63.92 \\ 
RIFE \citep{rife} & 19.54 & 0.4546 & 0.2954 & 57.68 & 25.73 & 0.7769 & \textbf{0.1564} & \textbf{42.33} \\ 
FILM \citep{reda2022film} & 19.75 & 0.4718 & 0.3048 & 68.88 & 24.96 & 0.7869 & 0.162 & 54.98 \\
LDMVFI \citep{danier2023ldmvfi} & 19.07 &	0.4175 &	\textbf{0.2765} & 56.28 & 24.53 & 0.7712 &\textbf{ 0.1564} & 42.96 \\
\hline
\textbf{VIDIM (ours)} & 18.73 & 0.4221 & 0.2986 & \textbf{ 53.38} & 22.88 & 0.688 & 0.1768 & 53.71 \\
\hline
\end{tabular}
\caption{Comparison between different video interpolation baselines and VIDIM on reconstruction and generative metrics, evaluating only the middle frame out of all 7 generated frames. VIDIM samples were obtained from our best cascade with guidance weight 2.0. Note that under this setting, it does not make sense to report FVD scores.}
\label{tab:sota-1}
\vspace{-2ex}
\end{table*}

\begin{table*}[t]
\small
\renewcommand\tabcolsep{5.0pt}
\centering
\begin{tabular}[t]{lccccc|lcccc}
&  \multicolumn{5}{c}{\textbf{Davis-7}}{\hskip 0.5cm} & \multicolumn{5}{c}{\textbf{UCF101-7}}{\hskip 0.25cm}  \\
\cline{2-11}
  & PSNR$\uparrow$ & SSIM$\uparrow$ & LPIPS$\downarrow$ & FID$\downarrow$ & FVD$\downarrow$ & PSNR$\uparrow$ & SSIM$\uparrow$ & LPIPS$\downarrow$ & FID$\downarrow$ & FVD$\downarrow$ \\ 
\hline
AMT \citep{amt} & \textbf{21.09} & \textbf{0.5443} & \textbf{0.254} & 34.65 & 234.5 & \textbf{26.06} & \textbf{0.8139} & 0.1442 & 31.6 & 344.5 \\ 
RIFE \citep{rife} & 20.48 & 0.5112 & 0.258 & 23.98 & 240.04 & 25.73 & 0.804 & 0.1359 & 18.72 & 323.8 \\ 
FILM \citep{reda2022film} & 20.71 & 0.5282 & 0.2707 & 30.16 & 214.8 & 25.9 & 0.8118 & 0.1373 & 26.06 & 328.2 \\
LDMVFI \citep{danier2023ldmvfi} & 19.98 & 0.4794 & 0.2764 & \textbf{22.1} & 245.02 & 25.57 & 0.8006 & \textbf{0.1356} & \textbf{18.09} & 316.3 \\
\hline
\textbf{VIDIM (ours)} & 19.62 & 0.4709 & 0.2578 & 28.06 & \textbf{199.32} & 24.07 & 0.7817 & 0.1495 & 34.48 & \textbf{278}\\
\hline
\end{tabular}
\caption{Comparison between different video interpolation baselines and VIDIM on reconstruction and generative metrics, evaluating all 7 generated frames. VIDIM samples were obtained from our best cascade with guidance weight 2.0. Note these numbers (especially FID scores) are not comparable to those in \ref{tab:sota-1} as the number of samples differs (here we use 7x as many images per set).}
\label{tab:sota-7}
\vspace{-2ex}
\end{table*}

We now provide a brief overview of the training objective formulation we utilize for VIDIM. We begin with the formulation of \citet{kingma2021variational}, where we use the simpler continuous-time objective (as opposed to that of \citet{ho2020denoising}) for learning a data distribution $p(\vect{x} | \vect{c})$, where $\vect{c}$ are the start and end frames and $\vect{x}$ are the middle frames. We define a forward process at every possible log signal-to-noise-ratio $\lambda$ (a.k.a. ``log-SNR'') in the usual manner via
\begin{equation}
    q(\vect{z}_\lambda | \vect{x}) = \mathcal{N}\left(\vect{z}_\lambda;\alpha_\lambda \vect{x}, \sigma_\lambda^2 \vect{I} \right)
\end{equation}
where $\alpha_\lambda = \sqrt{\mathrm{sigmoid}(\lambda)}$ and $\sigma_\lambda = \sqrt{\mathrm{sigmoid}(-\lambda)}$. Our training objective is then
\begin{equation}
    \mathop{\mathbb{E}}_{
        \tiny
        \substack{
        \vect{x,c}\sim p(\vect{x,c}) \\
        t\sim U(0,1) \\
        \vect{z}_{\lambda_t} \sim q(\vect{z}_{\lambda_t}|\vect{x})
        }
    }
    e^{\frac{\lambda_t}{2}} \left\|\alpha_{\lambda_t} \vect{z}_{\lambda_t} - \sigma_{\lambda_t} \vect{v}_\theta(\vect{z}_{\lambda_t} | \vect{c}) - \vect{x}  \right\|_1
\end{equation}
Note that, following \citet{saharia2022palette}, we use the L1 loss as we found early on that it helps produce better high-frequency details in samples compared to the standard L2 loss. We use a cosine log-SNR schedule $\lambda_t$ following \citet{kingma2021variational}, with maximum log-SNR of 20 at $t=0$ and minimum log-SNR of -20 at $t=1$. Note how this objective is equivalent to the re-weighted ELBO objective from \citet{ho2020denoising} in the sense that the loss is a norm between the predicted and actual noise (hence the $e^{\frac{\lambda_t}{2}}$ factor), but using the ``v-parametrization'' from \citet{salimans2022progressive} rather than predicting the added noise directly. This is well-known to improve training stability of diffusion models.

\section{Experiments}
\label{sec:experiments}

We now present our main experiments and results. All base models were trained to generate 7 64x64 frames in between the start and end 64x64 frames, and the super-resolution models condition on the original 256x256 start and end frames to upsample the 7 input frames at resolution 64x64.

In order to leverage a large-scale video dataset for these tasks, we train all VIDIM models on a mixture of the publicly available WebVid dataset \citep{Bain21} and other internal video datasets. In order to handle cases with large and more difficult motion, our input pipelines take bursts of 32 contiguous frames from the original videos and evenly subsample 9 frames so some frames are skipped. To additionally reduce the number of cuts and other examples that are undesirable for video interpolation, we follow the motion bracketing procedure employed by FILM \citep{reda2022film}.

\subsection{Training and architecture hyperparameters}

We trained all VIDIM models with the Adam optimizer \citep{kingma2014adam} with a learning rate of 5e-4 (with linear warm-up for the first 10,000 steps) and $\beta_1=.9, \beta_2=.999$, gradient clipping at norm 1, and maintaining an EMA of the model parameters with decay rate .9999 following \citet{ho2020denoising}. To make our ablation studies fair, \textit{all} base models were trained for 500k steps and all super-resolution models were trained for 200k steps. All super-resolution models were trained with noise conditioning augmentation \citep{ho2022cascaded} on the low-resolution frames, where we re-use the noise schedule and add noise to these frames with $t \in U(0, 0.5)$ for each training example.

We additionally study different parameter counts for our VIDIM models in \cref{tab:model_scaling_metrics}. In these experiments, we scale up the number of parameters exclusively by changing the hidden size (a.k.a. number of channels) in the \textit{last} UNet resolution (16x16 for both the base and super-resolution models). All other UNet resolutions have the same hyperparameters across experiments: the first resolution always has 128 channels and 2 subblocks, each subblock having a convolutional block and a temporal attention block with one attention head. Middle resolutions always have 256 channels and 4 subblocks, with each subblock's temporal attention block having two attention heads. The last 16x16 block has 8 subblocks, and additionally includes a self-attention block shared over frames before each temporal attention block. We avoid self-attention at other resolutions as it is too computationally expensive. We use dropout \citep{srivastava2014dropout} at all UNet resolutions with resolutions up to 64x64 but not beyond, following \citet{hoogeboom2023simple}.

\subsection{Comparisons with prior work}
\label{subsec:sota_comparison}
We evaluate VIDIM on several reconstruction and generative metrics, and compare against prior methods on video interpolation by running these models ourselves on the same benchmarks. Specifically, we create subsets of the Davis \citep{davis} and UCF101 \citep{soomro2012ucf101} datasets of 400 videos with examples that contain large and ambiguous motion and consisting of 9 frames per video. It should be emphasized that our numbers cannot be directly to prior work also due to the fact that diffusion models operate at a fixed resolution. Thus, in order to ensure that we are accurately representing all the prior work we consider, we run all the baselines ourselves on said benchmarks. We refer the reader to our Supplementary Material and website where we release our curated evaluation datasets (dubbed Davis-7 and UCF101-7) for future work and the general public, as well as playable video samples.

For each dataset, we report the following reconstruction-based metrics: peak-signal-to-noise-ratio (PSNR), structural similarity (SSIM) \citep{wang2004image}, and LPIPS \citep{zhang2018unreasonable}. We additionally report more popular metrics for generative models (specifically, FID \citep{heusel2017gans} and FVD \citep{unterthiner2018towards}), which, unlike reconstruction metrics, do not penalize \textit{plausible} extrapolations that differ from the ground truth. It is well-known that generative models should not be expected to achieve the best scores in reconstruction-based metrics \citep{sahak2023denoising_sr3plus,watson2022novel}. In fact, it has been consistently shown in prior work that blurrier images tend to score higher in reconstruction metrics despite being rated as worse by human observers \citep{chen2018fsrnet,dahl2017pixel,menon2020pulse,saharia2022image}. We nevertheless report said reconstruction metrics for both completeness and utility to compare different configurations of the same model, which we do find provides useful signal for development.

We additionally create a second set of results where we only evaluate the middle frame, similarly to prior work on video interpolation, as we find that otherwise the evaluation is skewed towards better frames: most frames are too close to the start and end frames and most methods do a good job on these, while it is clear that the middle frame is usually the most difficult. For these sets of results, we must exclude FVD as the I3D network requires a video with 9 frames as input. We show results of evaluating all 7 generated frames in \cref{tab:sota-7}, and results of evaluating only the middle frame as in \cref{tab:sota-1}. All samples were obtained using the ancestral sampler proposed by \citet{ho2020denoising} with 256 denoising steps per diffusion model.

Our quantitative evaluation demonstrates the superiority of VIDIM compared to other baselines in most generative metrics. While there is one exception, namely, that RIFE \citep{rife} achieves a better FID score than our cascaded model when evaluated over all 7 frames, this is not the case when considering only middle frame. The baseline methods will not have a difficult job with frames that are very close to the input frames, skewing the evaluation, while they clearly perform worse in the more difficult frames. This is also qualitatively apparent in our samples in \cref{fig:teaser} and in our samples in the Supplementary Material and website, especially in the cases with the largest amounts of motion. It is also noteworthy that VIDIM is always superior in FVD scores by a significant margin, showing that the aforementioned baselines produce much less natural-looking videos. This is an important quantity to consider, as FID does not consider any aspects of temporal consistency and only considers each frame individually. Notably, the RIFE baseline that achieves the best FID scores when considering all output frames, actually has the worst FVD scores and temporal consistency.

\begin{figure}[t]
    \includegraphics[width=\linewidth]{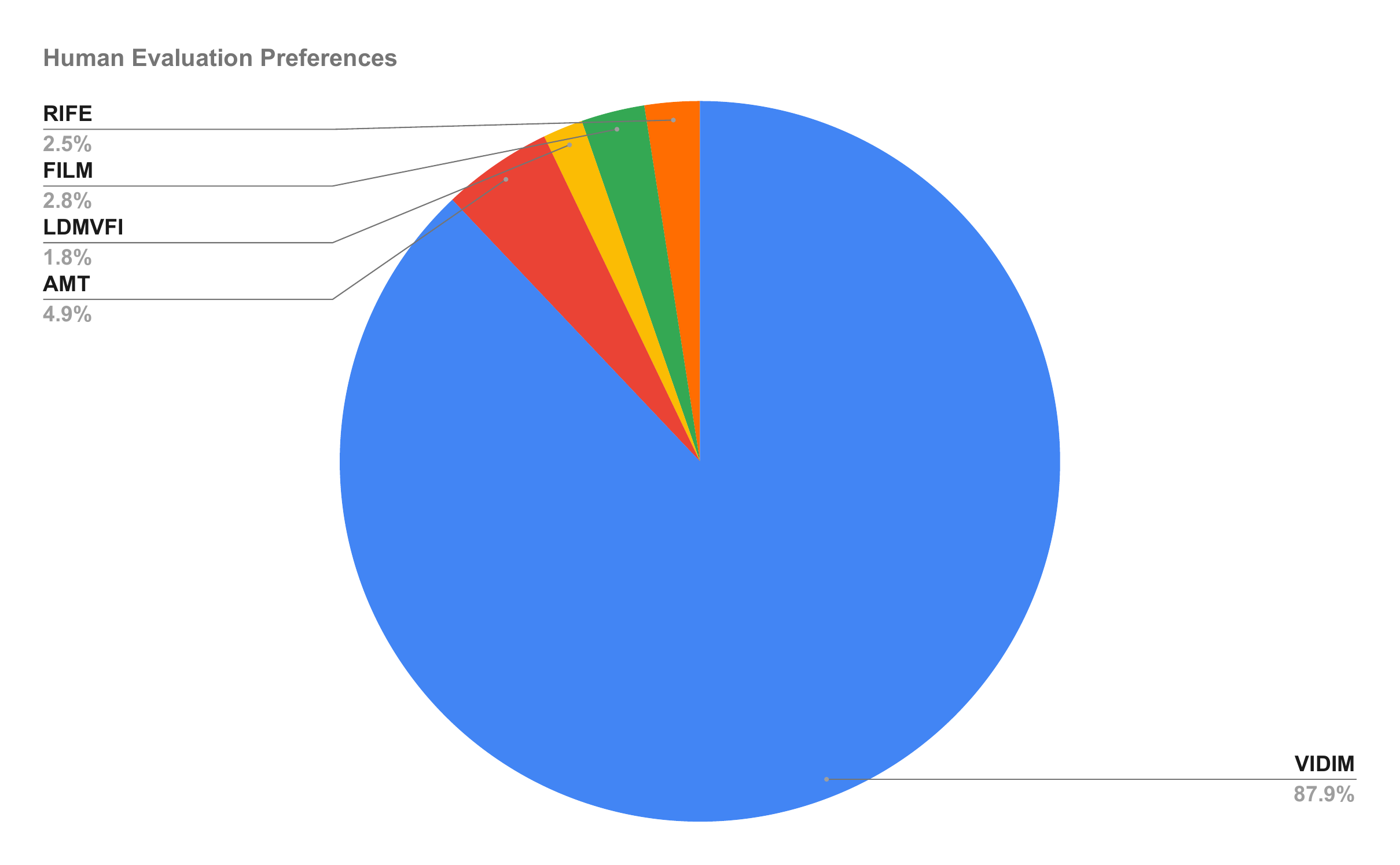}
    \caption{Human evaluation results on Davis-7, showing how often VIDIM and each baseline was preferred by human raters.}
    \label{fig:ratings_plot}
    \vspace{-2ex}
\end{figure}

\begin{figure*}[t!]
    \centering
    \includegraphics[width=\textwidth]{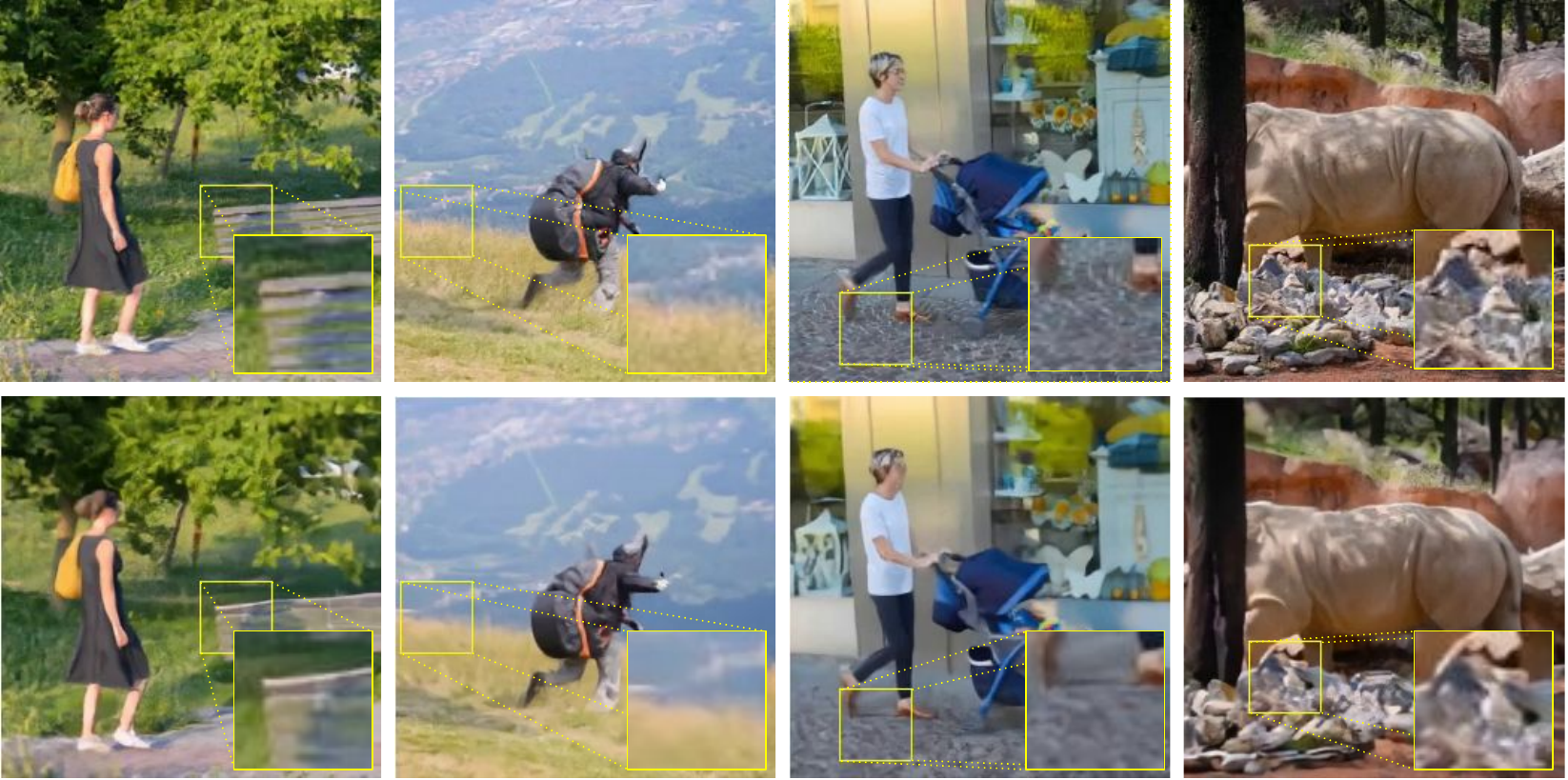}
    \caption{Sample comparison between our VIDIM medium super-resolution model (top) and an identically trained baseline minus high-resolution frame conditioning.}
    \label{fig:high_res_cond}
    \vspace{-1ex}
\end{figure*}

\subsection{Human evaluation}

To evaluate our method qualitatively, we additionally conducted a user-study where participants were shown video quadruplets playing side-by-side, each generated using the frame interpolation methods AMT, RIFE, FILM, LDMVFI and VIDIM (ours), from the same input frame pair. Users were shown up to 400 video quadruplet examples, in random order, and were asked to choose which of the four videos looks most realistic. The order used to layout the videos side-by-side in each example screen was also randomized. We evaluate on the Davis benchmark we used for all other evaluations in the paper, which contains both very challenging interpolation examples with ambiguous motion and difficult dis-occlusions, and very easy examples showing very little motion. The study involved 31 participants, and resulted in an aggregate of 1334 video quintuplet ratings. Each of the 400 examples was rated at least by one participant. Results in \cref{fig:ratings_plot} show that VIDIM samples are very strongly preferred by human observers. Moreover, if we normalize by the number of times each individual example was rated, results change by at most by 0.3\%, i.e., different participants tend to make the same choices.

\subsection{Ablation study on start+end frame conditioning}
\label{subsec:ablation_highresframe}
We now show the importance of explicitly training both the base and super-resolution models to be conditional on the input frames. To create a fair comparison, we train a super-resolution diffusion model that generates all 9 high-resolution frames of a video at training time, otherwise with all hyperparameters kept identical (including the number of training steps), and generate conditional samples via imputation: at every denoising step, we replace the start and end frame of the predicted $\vect{x}$ with the conditioning frames, add noise again and repeat. While this has been explored in prior work \citep{ho2022video}, our key hypothesis is that without the frame conditioning we propose, the training task becomes significantly more difficult. To further strengthen this baseline, we additionally try \textit{reconstruction guidance} following \citet{ho2022video}, which has been shown to improve sample quality by creating a similar effect to classifier-free guidance. Specifically, reconstruction guidance amounts to using the following prediction for $\vect{x}$ at every denoising step:
\begin{equation}
    \Hat{\vect{x}}_\textrm{guided} = \Hat{\vect{x}}_\textrm{inpaint} - (w - 1) \frac{\alpha_t}{2} \nabla_{\vect{z}_t}\left\| \vect{c} - \Hat{\vect{c}} \right\|_2^2
\end{equation}

where $w$ is the guidance weight (so $w=1$ corresponds to standard inpainting, hence the subtraction by 1, but higher values of $w$ entail more guidance), and $\Hat{\vect{c}}$ are the \textit{predicted} start and end frames that in standard inpainting we simply throw away.

\begin{figure}[b]
    \includegraphics[width=\linewidth]{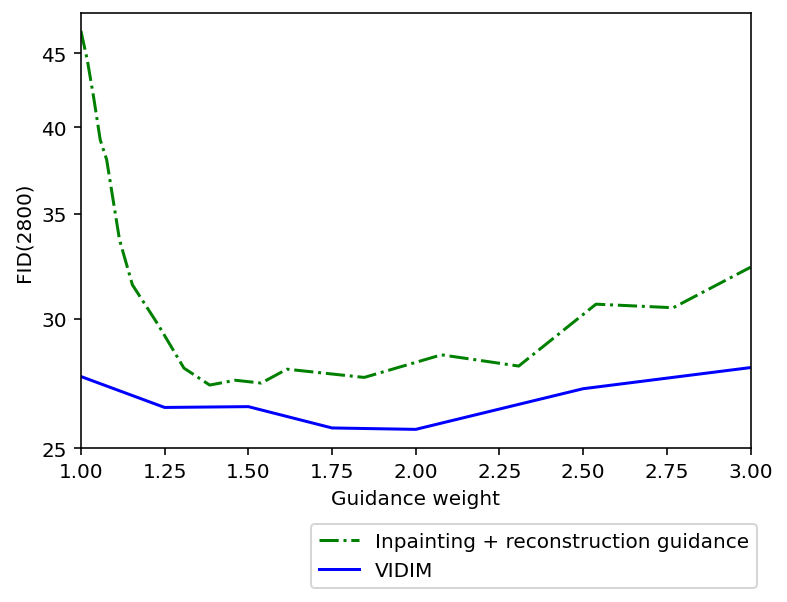}
    \caption{FID scores comparison between VIDIM and an inpainting baseline model at different guidance and reconstruction guidance weights, respectively. Note that the \textit{reconstruction} guidance weights (x-axis) for the baseline are re-scaled via $f(w)=(w-1)/13 + 1$ to more easily compare scores at the optimal region to VIDIM; the true range for the baseline guidance weights is from 1 to 27. The baseline model achieves an FID score of 60.11.}
    \label{fig:fidplot}
    \vspace{-1ex}
\end{figure}

\begin{table*}
\small
\renewcommand\tabcolsep{3.0pt}
\centering
\begin{tabular}{lccccc|lcccc}
&  \multicolumn{5}{c}{\textbf{Davis-7}}{\hskip 0.5cm} & \multicolumn{5}{c}{\textbf{UCF101-7}}{\hskip 0.25cm}  \\
\cline{2-11}
  & PSNR$\uparrow$ & SSIM$\uparrow$ & LPIPS$\downarrow$ & FID$\downarrow$ & FVD$\downarrow$ & PSNR$\uparrow$ & SSIM$\uparrow$ & LPIPS$\downarrow$ & FID$\downarrow$ & FVD$\downarrow$ \\ 
\hline
Base(lg) & 24.06 & 0.6987 & 0.094 & 21.15 & 116.42 & 23.04 & 0.6967 & 0.0848 & 24.39 & 194.6 \\
Base(md) & 22.89 & 0.6529 & 0.1108 & 22.59 & 116.48 & 23.04 & 0.6942 & 0.0857 & 25.34 & 198.0 \\
\hdashline
SSR(lg) & 27.76 & 0.7825 & 0.1468 & 23.94 & 132.68 & 31.49 & 0.8216 & 0.1554 & 33.84 & 168.368 \\
SSR(md) & 28.22 & 0.7976 & 0.1309 & 22.11 & 128.6 & 31.48 & 0.8288 & 0.1377 & 32.29 & 145.2\\
\hline
Base(lg) + SSR(lg) & 19.62 & \textbf{0.4709} & \textbf{0.2578} & 28.06 & 199.32 & \textbf{24.07} & \textbf{0.7187} & \textbf{0.1495} & \textbf{34.48} & 278  \\ 
Base(lg) + SSR(md) & \textbf{20.11} & 0.4632 & 0.3042 & 38.78 & \textbf{196.64} & 22.53 & 0.6775 & 0.2485 & 40.44 & \textbf{263.31}\\ 
Base(md) + SSR(lg) & 19.49 & 0.4481 & 0.269 & \textbf{26.58} & 217.14 & 24.1 & 0.7168 & 0.1507 & 35.88 & 280.8 \\
Base(md) + SSR(md) & 19.52 & 0.4327 & 0.3167 & 39.17 & 219.05 & 22.97 & 0.675 & 0.2513 & 42.24 & 279.83 \\
\hline
\end{tabular}
\caption{Comparison between different model sizes to illustrate the scalability of VIDIM. We compare the base and spatial super-resolution (SSR) separately and as a cascade. For each case, we consider two variants, a large model \textit{lg} and a medium model \textit{md}. Note that isolated models should only be compared to each other, as for the base model the ground truth is at a lower resolution, and the super-resolution models in isolation are conditioned on ground truth low-resolution frames.}
\label{tab:model_scaling_metrics}
\vspace{-4ex}
\end{table*}

Importantly, we note that for this to be a fair comparison, we only evaluate the baseline on the 7 middle frames, i.e. we explicitly discard the first and last frame. This is of paramount importance to not evaluate on any ground truth samples and to make the FID scores comparable (so they use the same number of samples). Results are included in \cref{fig:fidplot}. The baseline model achieves an FID score of 60.11 when disabling inpainting, i.w., when it has no access to any original high-resolution frames.

As we hypothesized, we find that conditioning on the high-resolution frames makes a significant difference in the quality of the results. Having access to sharp, small text helps the network preserve its legibility across the frames it generates. Even larger text, facial features, texture details, etc. can be botched without access to the high-resolution input frames. We provide a qualitative example in \ref{fig:high_res_cond}. Even without CFG, VIDIM models explicitly trained with the conditioning start and end frames achieve much better FID scores than the reconstruction guidance baselines. Interestingly, the range of ``good'' reconstruction guidance weights is quite different to CFG weights. Qualitatively, we find that it is key to use \textit{some} amount of CFG, but at CFG weights of around 4.0 and above, we begin noticing significant color artifacts.

\subsection{Scalability of VIDIM}

Finally, we also study the effect of scaling up the number of parameters of VIDIM models. As briefly mentioned in \cref{subsec:model_architecture}, we only change the hidden size (a.k.a. number of channels) of the \textit{last} UNet resolution to maximize memory savings. Because higher resolutions will have more activations \textit{per-parameter}, increasing the width of these layers is detrimental to peak accelerator memory usage as activations must be stored for backpropagation at training time. We thus increase the hidden size of the base model only at the 16x16 resolution from 1024 to 1792 and the number of attention heads from 8 to 14, making the parameter count change from 441M to 1.6B. For our super-resolution model, the hidden size at the 16x16 resolution was increased from 1024 to 1536 and the number of attention heads from 8 to 12, making the parameter count change from 644M to 1.01B. In order to avoid memory padding in the accelerator, the number of attention heads is always set such that the per-head hidden dimension is 128. With the use of ZeRO sharding \citep{rajbhandari2020zero}, both our medium and large models can be trained on accelerators with as little as 16GB of memory per chip at batch size 1 per chip, and there is enough remaining memory to maintain a ZeRO-sharded copy of the gradients to use microbatching and train on larger batch sizes with the same amount of accelerators. All our ``medium'' models are trained with a  batch size of 256, and our ``large'' models were trained with twice the data, i.e., at batch size 512, for the same number of training steps as their corresponding medium-sized models. Results are included in \cref{tab:model_scaling_metrics}.

Our quantitative results show the ability of VIDIM to favorably scale with more parameters and training: a larger base model is essential to not produce severe artifacts that will be amplified, and a larger super-resolution model is essential for sharpness in regions with the most amount of motion. Surprisingly, comparing the super-resolution models in isolation, the medium model achieves better quantitative results; however, using the large super-resolution is the most essential component to achieve low FID scores when sampling from the full cascade. Qualitatively, we see a clear difference in the samples; the large super-resolution model samples look sharper and have noticeably less artifacts. We hypothesize that the large super-resolution model has more capacity to hallucinate missing details and might be more robust than its medium counterpart to artifacts from the base model.

\section{Discussion and Future Work}

Through qualitative, quantitative and human evaluation, we show that our simple VIDIM models and architectures are capable state-of-the-art video interpolation, especially for large and ambiguous motion. Key components for good sample quality include explicit frame conditioning at training time and the use of classifier-free guidance. Still, several directions should be explored further. VIDIM-like models could be used for frame expansion, video restoration, among other tasks. Additionally, key problems remain to make these models maximally useful, including generating videos at arbitrary aspect ratios, and further improving the quality of super-resolution models. We also expect our architectures, framework, and demonstration of the effectiveness of image-guided conditioning will be generally useful for the community in other video generation tasks such as extrapolation, text-guided generation, and others.

{
    \bibliographystyle{ieeenat_fullname}
    \bibliography{main}
}

\clearpage
\setcounter{page}{1}
\maketitlesupplementary

\begin{figure}[t]
    \includegraphics[width=\linewidth]{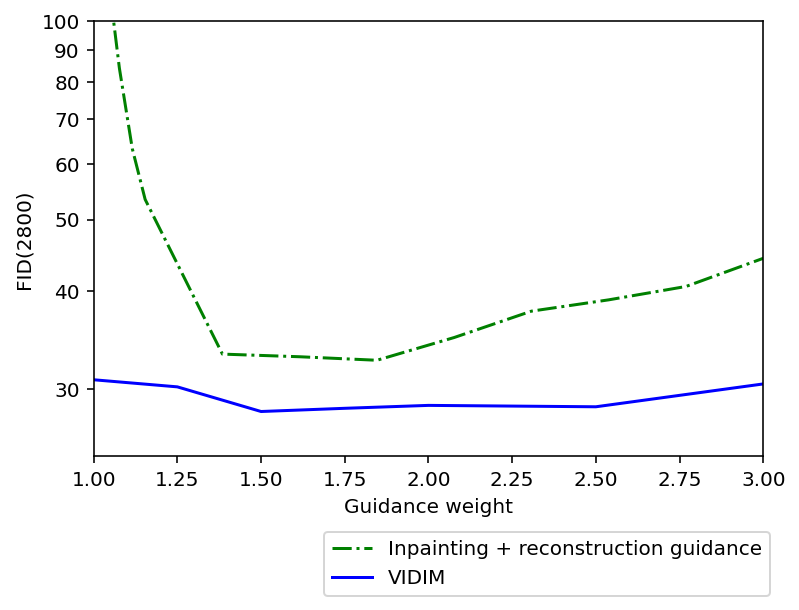}
    \caption{FID scores comparison between VIDIM and an inpainting baseline model at difference guidance and reconstruction guidance weights, respectively. Note that the \textit{reconstruction} guidance weights (x-axis) for the baseline are re-scaled via $f(w)=(w-1)/13 + 1$ to more easily compare scores at the optimal region to VIDIM; the true range for the baseline guidance weights is from 1 to 27.}
    \label{fig:fidplot_supp}
    \vspace{-2ex}
\end{figure}

\section{Supplementary website and more samples}
\label{sec:supp_website}
Please refer to our supplementary website \href{https://vidim-interpolation.github.io/}{https://vidim-interpolation.github.io/} for video outputs from VIDIM along with a comparison to the baseline methods studied in this work. We also provide downloadable zip files to the Davis-7 and UCF101-7 datasets we used for benchmarking as described in \cref{subsec:sota_comparison}.

We present some more qualitative comparisons against the baselines in \cref{fig:qr_supplementary}. As discussed in \cref{subsec:sota_comparison} these results demonstrate our method's ability to handle large, ambiguous motion between the start and end frames. The baselines, in comparison, tend to generate blurry and unnatural frames. We strongly encourage the reader to visit our \href{https://vidim-interpolation.github.io/}{Supplementary Website} to better appreciate the temporal dynamics present in the generated videos.

\section{Additional ablation studies}
In \cref{subsec:ablation_highresframe} we study the importance of explicitly training the super-resolution diffusion model to be conditional on the start and end frames. For the sake of completeness we also demonstrate the impact of this frame conditioning on the base diffusion model. Similar to \cref{subsec:ablation_highresframe} we  create a strong baseline by evaluating inpainting with reconstruction guidance as proposed in \cite{ho2022video}. We compare against the ``medium'' VIDIM model with 441M parameters. As shown in \cref{fig:fidplot_supp}, peak FID scores with reconstruction guidance are still worse than VIDIM despite both models being trained with identical parameter count, data and hyperparameters.

\begin{figure}[t]
    \includegraphics[width=\linewidth]{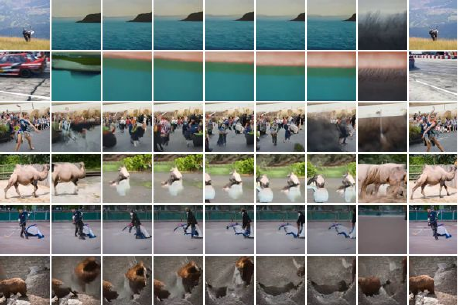}
    \caption{Frames generated via in-painting but with reconstruction guidance weight set to 1.0. The generated frames are not temporally consistent in the absence of reconstruction guidance. We roll out 9 frames (left to right) for six different videos (top to bottom).}
    \label{fig:supp_fig_no_cond_frames}
    \vspace{-1ex}
\end{figure}

In \cref{fig:supp_fig_no_cond_frames} we show the frames generated when reconstruction guidance weight is set to 1 in the inpainting + reconstruction baseline as described in the preceding paragraph. We found that without enough reconstruction guidance, standard in-painting cannot even produce consistent frames. 

\begin{figure*}
    \centering
     \subfloat{
        \includegraphics[width=0.85\textwidth]{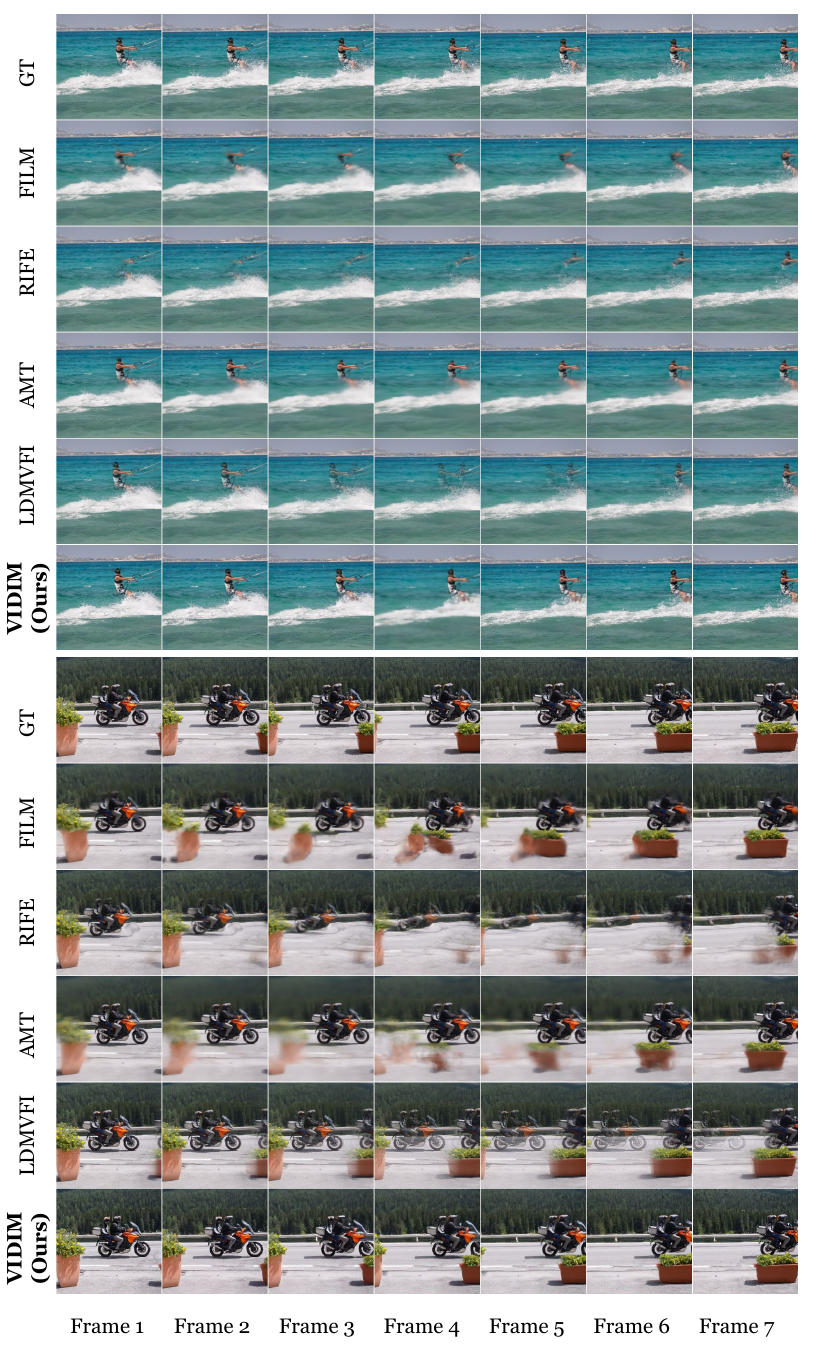}
    }
    \end{figure*}
\begin{figure*}
    \ContinuedFloat
    \centering
     \subfloat{
        \includegraphics[width=0.85\textwidth]{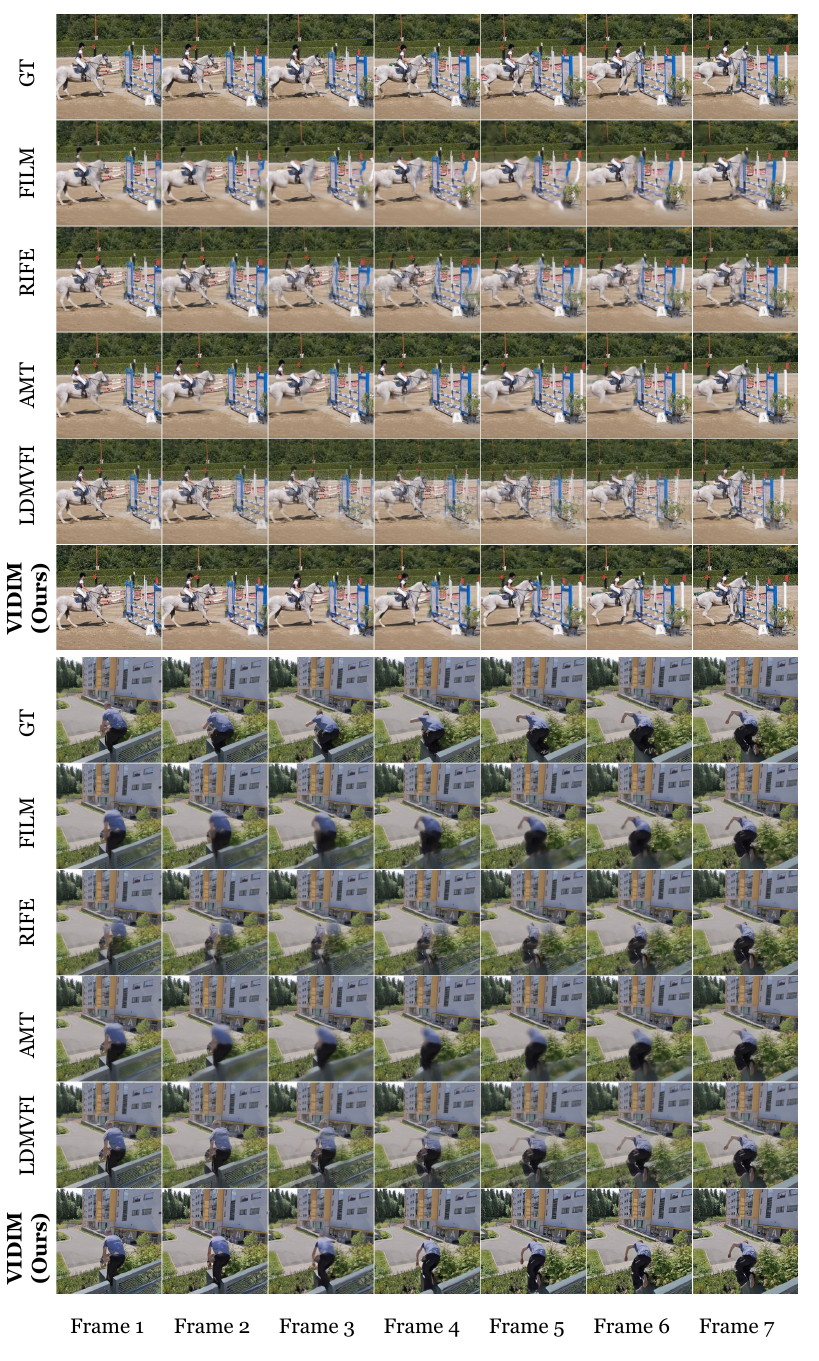}
    }
    \caption{Examples from the DAVIS-9 dataset showing the results from VIDIM compared to the baseline methods.}
    \label{fig:qr_supplementary}
    \vspace{4ex}
\end{figure*}

\end{document}